# Psychological Effects of Cultural Upheavals from Millions of Song Lyrics Over 100 Years

David M. Markowitz[1]

**Affiliation**

[1] Department of Communication, Michigan State University, East Lansing, MI 48824

## Abstract

Cultural upheavals impact many aspects of social life, and many studies have investigated their impact on language patterns. However, few investigations have isolated the impact of upheavals on individuals at scale in popular media. The current work evaluated millions of song lyrics spanning more than a century in search of within-artist and between-artist signals of distress from the Vietnam War, the terrorist attacks of 9/11, and COVID-19. Compared to a five-year baseline, rates of self-references — a marker of psychological distancing — were significantly reduced after the Vietnam War and September 11th. Cognitive processing terms were elevated post-upheaval vs. pre-upheaval, which indicated artists' increased attempts to make meaning from such massive disruptions. Content patterns corroborated these findings as artists wrote more about "life and freedom" (societal conditions) and less about "courtship and nightlife" (interpersonal connection) following the upheavals. Cultural upheavals modify individual and collective verbal behavior, demonstrating their far-reaching impact on society.

**Psychological Effects of Cultural Upheavals from Millions of Song Lyrics Over 100 Years**

The effects of psychological upheavals are pervasive and unrelenting. For example, people who face distressing life events like war, romantic breakups, or assault often report physical effects from such massive disruptions (Afari et al., 2014; Garfin et al., 2018), psychological impacts including depression and anxiety (Haehner et al., 2024; Kessler, 1997), and social consequences like a disinterest in engaging with others (Sandi & Haller, 2015). Upheavals not only impact individuals, but they can also impact collective thoughts, feelings, and behaviors as well (Chang, 2017; Hirschberger, 2018). The murder of George Floyd, for instance, negatively impacted the psychological well-being of the Black community in the US (Eichstaedt et al., 2021), COVID-19 continues to be a quality-of-life and financial stressor for the elderly (Iob et al., 2022), and events like the Holocaust have shaped the well-being and perceived safety of Jewish people generations into the future (Dashorst et al., 2019). Together, cultural upheavals can have both short- and long-term impacts on the communities they impact (Hirschberger, 2018; Nagata et al., 2024), making it important to understand the breadth and gravity of their consequences over time, and how they modify various forms of human behavior.

One way to understand how societal and cultural upheavals impact human behavior is to evaluate the words people use before and after such events. Evidence in clinical psychology, for example, has long demonstrated that traumatic experiences are reflected in how people use language and organize their thoughts. Narratives of traumatic experiences are often dominated by emotional and sensory details relative to other autobiographical accounts (Crespo & Fernández-Lansac, 2016), and as people work through their upheavals, more organized descriptions of the upheavals tend to be associated with a reduction in trauma-related symptoms (Foa et al., 1995). Language shifts from before to after an upheaval can therefore reveal changes in emotion (Pennebaker, 1997), social orientation (Badal et al., 2021), and cognitive processing (Booker et al., 2018; Kleim et al., 2018), providing linguistic markers of peoples' underlying psychological states (Boyd & Markowitz, 2024; Boyd & Schwartz, 2021). Crucially, these patterns extend to upheavals in the wild. Following the September 11 terrorist attacks, for example, bloggers wrote with more negative emotions and a greater rate of psychological distancing (e.g., a reduced rate of self-references) relative to baseline (Cohn et al., 2004), with many verbal patterns failing to return to baseline weeks later. In sum, a wealth of evidence suggests language offers a psychological record of how people respond to major societal and cultural disruptions over time.

The current work draws on a long tradition of using language patterns to study cultural upheavals, asking how popular music reflects musicians' meaning-making after major disruptions. Musicians and other artists are often cultural interpreters and social commentators, communicating their own beliefs in song but also reflecting on the pulse of society and culture at the time their songs are written (Blau, 1988; Chye & Kong, 1996; Tochka, 2023). Therefore, much like a time capsule that contains thoughts, feelings, and reflections that are frozen in time (Markowitz et al., 2025), song lyrics make for an essential data source to understand how personal, social, and cultural reflections appeared in words as a function of the context and conditions people were living in. Using millions of song lyrics, the present work evaluates how musicians and society processed three major societal and cultural events, including the Vietnam War, 9/11, and the COVID-19 pandemic.

What is unique about this investigation, relative to other studies evaluating the impact of upheavals on language use, is the availability of within-artist and between-artist song lyric data at scale. Within-artist trends measure how artists themselves responded to such upheavals and between-artist trends measure on average, how artists whose songs came after an upheaval differ linguistically from artists whose songs mostly came before an upheaval. The ability to evaluate such trends within-person, at the scale of millions of song lyrics over time, is rare and can help to describe the trajectory in recovery from societal and cultural disruptions.

**The Impact of Societal and Cultural Upheavals on Verbal Behavior**

Research in the psychology of language field often uses words as markers of attention to approximate mental processes of the individual mind (Boyd & Markowitz, 2024; Boyd & Schwartz, 2021; Pennebaker, 2011). Therefore, in the study of upheavals and psychological distress, scholars often try to evaluate where people focus their attention after a disruption relative to before a disruption using different verbal signals. The first set of signals include style words, or words that describe *how* a person is communicating instead of what they are communicating about. Style words include pronouns (e.g., self-references or "I"-words, collective references or "we"-words) and studies of personal upheavals have demonstrated that people often turn to others for social support and social sharing during such disturbances, which tends to be associated with an increased rate of collective-references (e.g., *our*, *us*) (Gortner & Pennebaker, 2003; Liehr et al., 2004; Mehl & Pennebaker, 2003; Rimé et al., 1998). When people experience shared societal or cultural upheavals, their communicative style tends to be less self-focused (Cohn et al., 2004; Stone & Pennebaker, 2002), presumably to manage an upheaval by removing the self from it for protection.

Other indicators of psychological processing during upheavals including cognitive dimensions as well (Davis & Nolen-Hoeksema, 2001). Prior work suggests people who experience a traumatic event or psychological disturbance often try to understand it and make meaning from the event (Cohn et al., 2004; Markowitz, 2022; Seraj et al., 2021). This meaning-making process is associated with an increase in cognitive processing terms (e.g., *because*, *should*), which indicate a person who is "working through" an unresolved psychological experience. People who were broken up and discussed their experience on reddit used their highest rate of cognitive processing terms just as the break-up occurred (Seraj et al., 2021), and bloggers after 9/11 were most unsettled days after the terrorist attack, returning to baseline approximately one week after (Cohn et al., 2004). Second, people who experience psychological traumas and upheavals tend to communicate in a more narrative-like and informal manner compared to an analytical and formal manner (Markowitz, 2022; Seraj et al., 2021). Prior work has applied an analytic thinking index (Pennebaker et al., 2014) — which reflects high rates of articles and prepositions, but low rates of storytelling words like pronouns — to a range of settings, with evidence generally suggesting that rates of analytic thinking decrease after a psychological disturbance compared to before (Monzani et al., 2021; Seraj et al., 2021). Scientists who wrote papers after the COVID-19 pandemic, for example, communicated in a more informal and less analytic manner compared to scientists who wrote papers pre-pandemic (Markowitz, 2022). Therefore, in general, upheavals tend to be associated with cognitive shifts, namely those that involve an increase in cognitive processing and a decrease in analytic thinking.

**The Current Paper**

The goal of the present study is to evaluate shifts in song lyrics as a function of societal and cultural upheavals. Three upheavals were evaluated: (1) the Vietnam War, (2) September 11th, and (3) The COVID-19 pandemic. These upheavals were chosen because they rank among some of the largest societal and psychological disruptions in history (Deane, 2016; Naveed et al., 2024) and they are temporally diverse enough to rule out time-dependent effects. These upheavals have also received significant attention in the social scientific literature (Bavel et al., 2020; Cohn et al., 2004; Rothbaum et al., 2001; Wang et al., 2025). Evaluating how they impact popular media like song lyrics demonstrates the reach of their consequences across many aspects of the human condition and the durability of such consequences.

It is also important to reflect on why song lyrics provide a unique and critical dataset to evaluate upheavals. First, song lyrics are psychological records that few popular media outlets can match at scale. The present corpus contains roughly 3 million songs released between 1903 and 2026, providing substantial coverage around each upheaval and enough power to detect small, but reliable shifts in language. Second, lyrics are nested within artists, and most artists release music repeatedly over time. This allows changes from cultural upheavals to be evaluated within-artist and between-artist. Within-artist effects capture whether the same artists changed how they wrote after an upheaval compared to before an upheaval, and between-artist effects capture whether the sample of artists shifted, on average, over time. It is often difficult to evaluate longitudinal within-person effects in observational data, making song lyrics a unique source of psychological information about individuals and aggregates. Finally, lyrics are unobtrusive "time capsules" that produce in-the-moment reflections of thoughts and feelings. They offer access to what it felt like to live through an upheaval before their individual and collective consequences were known.

Song lyric data have provided rich insights into a range of social and psychological dimensions in prior work (Berger & Packard, 2018; Brand et al., 2019; DeWall et al., 2011; Varnum et al., 2021), yet relatively little is known about how language patterns from song lyrics are modified by societal upheavals (Foramitti et al., 2025), whether such changes occur within the same artists, and how long any verbal shifts persist before returning to baseline. The current work addresses these questions by using more than a century of song lyrics to examine the immediate and long-term linguistic effects of major societal and cultural upheavals.

# Method

## Data Collection and Preprocessing

Over 32 million song lyrics were obtained from LRCLIB, an online database of publicly available song lyrics. The full catalogue of lyrics and metadata (e.g., artist, song title, album) was preprocessed in the following manner to create the paper's dataset. First, lyrics were scanned with language detection software (i.e., fastText's *lid.176.bin* model) to identify those in English and other languages (Joulin et al., 2016). Consistent with best practices (Boyd et al., 2022; Meier et al., 2020), non-English lyrics were translated to English using Meta's *nllb-200-distilled-600M* model (NLLB team et al., 2022) because language dimensions of interest required English texts.

Additional metadata from other sources were appended to the LRCLIB database to create the focal sample. Song-level metadata (e.g., International Standard Recording Code, song duration, release date, genre, and record label) was obtained from MusicBrainz's publicly available bulk data dumps (MusicBrainz, 2026). Each LRCLIB track was matched to a MusicBrainz song via normalized artist-name and track-title string matching, cross-validated against recording duration, and assigned to one of four confidence tiers (A through D) based on the closeness of the string and duration match: Tier A (string and duration match within ±2 seconds, with matching album name), Tier B (string and duration match within ±2 seconds), Tier C (string and duration match within ±5 seconds), and Tier D (string match only, with no MusicBrainz duration available for comparison). Songs with a string match but a duration disagreement exceeding 5 seconds were classified as weak matches and excluded from the final dataset. In total, 15,121,980 of 32,254,478 songs (46.9%) received a usable match (those from Tiers A-D were retained). Because LRCLIB issues a separate track record for each album, reissue, and compilation credit, records were then collapsed to one row per recording, yielding a total of 11,102,786 songs. Analyses required a release date and non-missing LIWC scores, leaving a final sample of 3,064,753 songs for the current paper.

**Automated Text Analysis**

All lyrics were processed with Linguistic Inquiry and Word Count (LIWC) (Boyd et al., 2022), a program that counts words as the percentage of the total word count per text against a dictionary of social (e.g., words about family), psychological (e.g., words about emotion), and part of speech dimensions (e.g., articles, pronouns). Four dimensions, including self-references (e.g., *I*, *me*), collective-references (e.g., *we*, *us*), analytic thinking (e.g., a standardized composite of eight style word categories, with high scores indicating formal and analytic writing relative to informal and narrative-like writing) (Markowitz, 2023a; Pennebaker et al., 2014), and cognitive processing terms (e.g., *because*, *should*) were evaluated based on their connection to prior research (Cohn et al., 2004; Markowitz, 2022; Mehl & Pennebaker, 2003; Seraj et al., 2021; Stone & Pennebaker, 2002). Song lyric excerpts and their associated LIWC scores for the key dimensions of interest are located in Table 1.

**Analytic Plan**

To evaluate how the communication patterns of song lyrics were modified by cultural upheavals, several analytic procedures were performed. Each cultural upheaval was operationalized as a single critical date with a symmetric five-year observation window. A five-year observation window was selected to capture a wide enough time-period to allow musicians to release songs before and after an upheaval (Fowler et al., 2013; Hendricks & Sorensen, 2009). Song release dates were available at day, month, or year levels of precision. A song was coded as “pre-upheaval” if its entire possible release interval fell before the upheaval date and “post-upheaval” if that interval fell entirely after it. Songs whose interval straddled the date could not be assigned to a period and were excluded. The three upheavals and their critical dates were the escalation of the Vietnam War with US ground troops in the region (date = March 8, 1965), the September 11 attacks (date = September 11, 2001), the onset of the COVID-19 pandemic (date = January 10, 2020, or when the World Health Organization first mentioned a novel coronavirus).

The post-upheaval indicator (e.g., 1 = song released after an upheaval, 0 = song released before an upheaval) was separated into within-artist and between-artist effects through group-mean centering consistent with prior work (Enders & Tofighi, 2007; Mundlak, 1978). For each artist, the mean of the post-upheaval indicator was calculated across their songs within the upheaval window, producing a proportion of that artist’s song catalog released after the upheaval. This artist mean was the between-artist predictor, and each song’s deviation from its artist’s mean served as the within-artist predictor (Curran & Bauer, 2011). Artists with songs only on one side of an upheaval have a within-artist value of zero and therefore contribute to the between-artist estimate but not the within-artist estimate. Within-artist effects ask if the same artists write differently after the event versus before it, and between-artist effects ask if artists whose songs mostly come after the event differ linguistically from artists whose songs mostly fall before it, on average.

Statistical models were fit using linear mixed models (Bates et al., 2015; Kuznetsova et al., 2020) with random intercepts for artist, primary song genre, and detected source language to control for content-based covariates. However, genre and language metadata were unavailable for some songs and rather than dropping these cases, unlabeled songs were assigned to an “unknown” level.

# Results

## Descriptive Patterns

Descriptive statistics and intercorrelations for the four key language variables are in Supplementary Table S1 and song counts by year in Supplementary Table S2.

The evidence in Figure 1 tracks patterns of self-references, collective-references, analytic thinking, and cognitive processing over time across all 3 million song lyrics. Consistent with prior work (DeWall et al., 2011), these trends suggest popular media reflects an increased self-focus ($r = 0.064$, $p < .001$) and collective focus ($r = 0.048$, $p < .001$) over time. Finally, there has been an increase in both cognitive processing terms ($r = 0.042$, $p < .001$) and a decrease in analytic thinking ($r = -0.036$, $p < .001$), despite the negative relationship between these two variables ($r = -0.477$, $p < .001$). These effect sizes are noticeably, and expectedly, small.

## Cultural Upheavals

**Vietnam War.** Within-artist effects revealed that compared to before the Vietnam War, self-references decreased ($t = -3.35$, $p = .001$) and rates of cognitive processing terms increased after the Vietnam War ($t = 3.07$, $p = .002$). Models approximating the relationship between post-upheaval and collective-references ($p = .427$) and analytic thinking ($p = .120$) were not statistically significant. The only statistically significant between-artist effect was related to cognitive processing terms ($t = 4.06$, $p < .001$), which suggests artists whose songs mostly came after the Vietnam War contained a higher rate of cognitive processing terms compared to artists whose songs mostly came before it.

**September 11$^{th}$.** Within-artist effects revealed that compared to before 9/11, self-references ($t = -7.38$, $p < .001$) and rates of analytic thinking ($t = -3.74$, $p < .001$) decreased after 9/11. Compared to before 9/11, rates of collective references ($t = 13.66$, $p < .001$) and cognitive processing terms ($t = 2.07$, $p = .039$) increased after 9/11. Between-artist effects were all positive and statistically significant (*p*s < .001) except for analytic thinking, where the relationship was negative and statistically significant ($p < .001$).

**COVID-19.** There was a marginally significant, negative relationship between post-upheaval time and self-references ($t = -1.94$, $p = .053$) and post-upheaval time and collective references ($t = -1.83$, $p = .067$) at the within-artist level. Rates of analytic thinking ($t = 6.93$, $p < .001$) and cognitive processing terms ($t = 7.92$, $p < .001$) were higher after versus before COVID-19 as well. All between-artist effects were significant with self-references and cognitive processing terms being higher post-COVID compared to pre-COVID (*p*s < .001), and collective references and analytic thinking being lower post-COVID compared to pre-COVID (*p*s < .001).[1]

All song and artist counts across individual cultural upheavals are represented in Supplementary Table S4.

## Exploratory Within-Person Changes

When do the psychological effects of cultural upheavals return to an artist's baseline? To address this question, a lagged analysis was estimated for each upheaval and each dependent variable. Songs were assigned to twelve-month bins according to their release date relative to the upheaval, spanning the five years before each event and then up to ten years after it (results using six-month bins are in Supplementary Figure S1 for completeness). Bins containing fewer than 200 songs were excluded out of caution that estimates would be difficult to interpret with fewer cases. Each dependent variable was regressed on bin indicators with fixed effects for artist, primary genre, and detected source language, with standard errors clustered by artist (Berge et al., 2026). Artist fixed effects isolate within-artist change, where each bin coefficient represents the estimated deviation in an artist's language patterns from their pre-upheaval baseline periods.

[1] Many lyrics in this database were translated from different languages to English and therefore, some translations may have missed nuances of original message. Therefore, using English-only cases, the models were rerun. Supplementary Table S3 reveals that the overwhelming majority of results replicated using English-only cases (e.g., 21 of 24 retained their direction and 20 of 24 their significance classification). Additional sensitivity checks are offered in the online supporting information.

The evidence in Figure 2 represents within-artist deviations from baseline over time across the three cultural upheavals in binned models. For the Vietnam War, within-artist rates of self-references declined steadily following the escalation and remained below baseline for over a decade while rates of analytic thinking rose over the same period. Collective-references and cognitive processing terms vacillated in both directions after the baseline period with few stable trends. For 9/11, self-references fell modestly within-artist and remained below baseline over the course of a decade, while collective-references rose steadily over time. Analytic thinking and cognitive processing terms hovered around baseline and were substantively unchanged over time. Finally, within-artist patterns for COVID-19 were less pronounced and more longitudinal in nature relative to the other upheavals. The most noticeable pattern was an increase in rates of analytic thinking over the course of several years compared to baseline. Therefore, these results suggest within-artist effects are apparent in how musicians communicate as a function of upheavals they and society experience in everyday life.

**Exploratory Content Effects**

To evaluate how these effects were robust to content patterns, the data were submitted to the Meaning Extraction Method (Chung & Pennebaker, 2008; Markowitz, 2021), a technique that clusters content words (e.g., nouns, verbs) through principal component analysis for the development of themes (see online supporting information for full details). The evidence in Table 3 suggests 7 themes were retained across all 3 million lyrics, including topics related to *courtship*, *devotion*, and *moral conflict*. Themes were saved as regression weights for use as controls in the prior mixed model calculations and also separately as dependent variables in the pre- versus post-upheaval models. See Supplementary Figure S2 for general trends over time for these themes.

The results were consistent when extracted themes were added as controls with two nuances (see Supplementary Table S5). First, Vietnam War models without thematic controls revealed a non-significant negative relationship between time post-upheaval and collective references for the between-artist effect ($p$ = .197), but with controls, the model obtained statistical significance ($p$ = .039). A similar pattern happened for COVID-19, where models without thematic controls revealed a marginally significant negative relationship between time post-upheaval and collective references for the within-artist effect ($p$ = .067), but with controls, the model obtained was statistically significant ($p$ = .037).

In predicting themes from post-upheaval (versus pre-upheaval) and random intercepts controls for artist, primary song genre, and detected source language, the within-artist results suggested "life and freedom" was greater post-upheaval relative to pre-upheaval, and "courtship and nightlife" was lower post-upheaval relative to pre-upheaval. The theme of "moral conflict" was greater post-upheaval compared to pre-upheaval in all upheavals as well, though the Vietnam estimate was not statistically significant (see Supplementary Table S6 for all results). In sum, seven themes were extracted from millions of song lyrics and themes related to freedom and moral conflict were often elevated post-upheavals relative to pre-upheavals.

## Discussion

The current work evaluated millions of song lyrics to uncover the psychological effects of three major upheavals in the US: (1) the Vietnam War, (2) the terrorist attacks of September 11$^{th}$, and (3) the COVID-19 pandemic. Within-artist effects revealed that compared to a five-year pre-upheaval period, rates of self-references were significantly reduced after the Vietnam War and September 11th in a post-upheaval five-year period (self-references were marginally reduced after COVID-19 relative to baseline as well). Patterns of cognitive processing terms, including words like *because*, *should*, or *ought*, were elevated post-upheaval relative to pre-upheaval, which indicate artists' increased attempts to work through and make meaning from massive societal and cultural disruptions. Findings for other language dimensions of interest (i.e., collective-references, analytic thinking) were more mixed across upheavals. Finally, longitudinal within-artist results were assessed to understand their trajectory and return-to-baseline potential. The reduction in self-references for the Vietnam War and September 11th remained below baseline for over a decade. Further, rates of collective-references remain elevated following September 11th approximately ten years later.

There are several contributions of this work. First, the within-artist and between-artist effects separate two processes that prior work on upheavals and language has conflated or ignored. Within-artist estimates ask whether the same writers changed from pre-upheaval to post-upheaval, and between-artist estimates ask whether the sample of artists changed over time. Importantly, in this dataset, they often operated in opposite directions. After 9/11, for example, self-references declined within-artist but increased between-artist, indicating that individual artists distanced the self at the same time that other artists, on average, were more self-focused. In another example, after COVID-19, certain within-artist effects were positive (e.g., higher rates of analytic thinking post-upheaval relative to pre-upheaval) and between-artist effects were negative. A design that does not isolate these effects would report a single estimate corresponding to neither process, and would have missed the differences entirely.

In addition to within- versus between-artist effects, it is also critical to evaluate the longitudinal return-to-baseline patterns for certain upheavals. For the Vietnam War, it was perhaps surprising to observe that the reduction in self-references required several years to become significantly different from baseline. Why did this delay in the psychological impact of such an upheaval occur? One explanation is that the date of the event in this analysis (March 8, 1965) was an escalation of the war and when US deployed its first ground troops in South Vietnam. The war did not officially end until April 30, 1975. Therefore, it is reasonable that because the war was enduring, it took some amount of time for the social, cultural, and psychological effects of such war to permeate popular media like song lyrics. The evidence in this paper suggests such effects were palpable in song lyrics approximately three-to-four years after the war's escalation and did not return to baseline over the course of a decade. A similar question pertains to September 11th, where the within-artist trajectory was more striking than the average between-artist effect. Cohn et al. (2004) found that some verbal markers returned to baseline within roughly two weeks of the attacks, while psychological distancing remained elevated and cognitive processing fell below baseline in an eight-week window. The present evidence indicates that within-artist collective-references remained elevated approximately a decade later, and that self-references remained below baseline over the same period (the self-references pattern is generally consistent with Cohn and colleagues). Therefore, despite lyrics being written and released with a time-lag relative to blog posts, both media types capture the social and psychological effects of upheavals. In other words, upheavals are encoded in our psychology regardless of the medium that is used to express and record them.

A third contribution concerns what artists wrote about. As controls, content results indicate that the style effects reported here are not a function of shifting topics (the within- and between-artist estimates were substantively unchanged when thematic content was included). As outcomes, content results indicated upheavals altered what musicians wrote about in ways that converge with the style results. Artists wrote more about “life and freedom” and less about “courtship and nightlife” following each event, with “moral conflict” being elevated post-upheaval during 9/11 and COVID-19 as well. This pattern, which reflects less romantic and social content and more existential and moral content, is generally consistent with the within-artist reduction in self-references and elevation of cognitive processing terms. One interpretation is that content about courtship tends to be interpersonal in nature (e.g., addressed from one person to another), whereas content about freedom and moral conflict tends to describe broader social or societal conditions. Such a shift is at least compatible with a reduction in self-references and more attempts at meaning-making (e.g., causal reasoning), though future work would be required to evaluate how directly connected these contentions are at scale.

Finally, this paper is a major extension of prior lyrics-as-data research, which have often evaluated psychological effects in lyrics from the most popular songs or artists (Berger & Packard, 2018; Foramitti et al., 2025), songs from a limited time period (DeWall et al., 2011), or lyrics from a limited number of songs (Varnum et al., 2021). The current work improves on all three fronts, using over 3 million songs between 1903 and 2026 with the ability to dissect within-artist patterns from more general between-artist patterns. Methodologically and theoretically, this work provides a more nuanced understanding of the lifecycle of psychological upheavals from the lens of song lyrics that other work has been unable to appreciate.

**Limitations and Future Directions**

The results in this study are not cause and effect, and should therefore be treated as correlational. Further, the effect sizes are notably small but consistent with those represented in the psychology of language field (Kern et al., 2014; Kramer et al., 2014; Markowitz, 2023b;

Shulman et al., 2024). These patterns were observable likely due to the scale and scope of the present study (e.g., over 3 million cases). It is also worth noting that while translations from non-English to English were required for the purposes of using the natural language processing tool in this study, there may be some nuance that is missed in the translation process. Prior work finds extremely high correlations between human- and machine-translated sentences for LIWC-based dimensions (average $r = .820$) (Windsor et al., 2019), but there is some small amount of error in machine translations. For millions of song lyrics, human translation was impossible, but this limitation is worth acknowledging.

The cultural upheavals in this study were also US-centric and future investigations should examine how more global events impacted language patterns in popular media. While this dataset contains multiple languages that were translated to English, future work might examine how nationality and distance to a cultural upheaval might modify the results.

**Table 1**

*Excerpts from the Dataset*

| Artist | Song | Dimension | Lyrical Excerpt | Excerpt Score |
|---|---|---|---|---|
| Prince | I Will | High self-references | I will walk this road, I will, I will<br>It's gonna be hard I know, but I will, I will<br>People come and they'll go, but I still, I still<br>Face up to the truth and just grow, I will, I will | 23.81% |
| Bing Crosby | Hello Dolly | Low self-references | You're lookin' swell, Dolly<br>I can tell, Dolly<br>You're still glowin', you're still crowin'<br>Yes, you're comin' on real strong | 5% |
| Earth, Wind & Fire | Evil | High collective-references | Beauty in our face, you see<br>Tryin' to hide all our misery<br>Our misery, our misery, our misery | 27.78% |
| Eric Clapton | Layla | Low collective-references | What will you do when you get lonely<br>And no one's waiting by your side?<br>You've been running and hiding much too long<br>You know it's just your foolish pride | 0% |
| Fleet Foxes | Sun It Rises | High analytic thinking | The sun rises<br>Over my head<br>In the morning<br>When I rise | 89.52 |
| Billy Joel | My Life | Low analytic thinking | I don't need you to worry for me, 'cause I'm alright<br>I don't want you to tell me it's time to come home<br>I don't care what you say anymore, this is my life<br>Go ahead with your own life, leave me alone | 1.00 |
| Neil Young | Change Your Mind | High cognitive processing | When you get weak, and you need to test your will<br>When life's complete, but there's something missing still<br>Distracting you from this must be the one you love<br>Must be the one whose magic touch can change your mind<br>Don't let another day go by without the magic touch | 26.0% |
| The Who | Won't Get Fooled Again | Low cognitive processing | I'll tip my hat to the new constitution<br>Take a bow for the new revolution<br>Smile and grin at the change all around me<br>Pick up my guitar and play<br>Just like yesterday<br>And I'll get on my knees and pray | 2.44% |

**Table 2**

*Linear Mixed Model Results*

| Upheaval | Dependent variable | Fixed effect | *B* | *SE* | *df* | *t* | *p* | Cohen's $f^2$ | Observations | Artists | Spanning | Informative | ICC |
|---|---|---|---|---|---|---|---|---|---|---|---|---|---|
| Vietnam War | Self-references | Intercept | 6.69 | 0.35 | 49.09 | 19.3 | < .001 | | 43714 | 5015 | 509 | 23779 | 0.086 |
| Vietnam War | Self-references | Within-artist | -0.27 | 0.08 | 40257.57 | -3.35 | .001 | 2.76E-04 | 43714 | 5015 | 509 | 23779 | 0.086 |
| Vietnam War | Self-references | Between-artist | -0.17 | 0.12 | 4008.53 | -1.41 | .159 | 4.52E-04 | 43714 | 5015 | 509 | 23779 | 0.086 |
| Vietnam War | Collective-references | Intercept | 1.11 | 0.1 | 38.51 | 10.93 | < .001 | | 43714 | 5015 | 509 | 23779 | 0.032 |
| Vietnam War | Collective-references | Within-artist | -0.02 | 0.03 | 40869.84 | -0.79 | .427 | 1.45E-05 | 43714 | 5015 | 509 | 23779 | 0.032 |
| Vietnam War | Collective-references | Between-artist | -0.05 | 0.04 | 2926.76 | -1.29 | .197 | 5.71E-04 | 43714 | 5015 | 509 | 23779 | 0.032 |
| Vietnam War | Analytic thinking | Intercept | 43.47 | 2.35 | 40.82 | 18.47 | < .001 | | 43714 | 5015 | 509 | 23779 | 0.121 |
| Vietnam War | Analytic thinking | Within-artist | 0.65 | 0.42 | 39967.11 | 1.56 | .120 | 6.06E-05 | 43714 | 5015 | 509 | 23779 | 0.121 |
| Vietnam War | Analytic thinking | Between-artist | -0.33 | 0.7 | 4543.87 | -0.47 | .636 | 4.93E-05 | 43714 | 5015 | 509 | 23779 | 0.121 |
| Vietnam War | Cognitive processes | Intercept | 6.81 | 0.32 | 44.88 | 21.14 | < .001 | | 43714 | 5015 | 509 | 23779 | 0.092 |
| Vietnam War | Cognitive processes | Within-artist | 0.24 | 0.08 | 40241.31 | 3.07 | .002 | 2.22E-04 | 43714 | 5015 | 509 | 23779 | 0.092 |
| Vietnam War | Cognitive processes | Between-artist | 0.49 | 0.12 | 4110.68 | 4.06 | < .001 | 4.22E-03 | 43714 | 5015 | 509 | 23779 | 0.092 |
| 9/11 | Self-references | Intercept | 6.72 | 0.22 | 78.44 | 30.6 | < .001 | | 597988 | 80845 | 10455 | 374465 | 0.142 |
| 9/11 | Self-references | Within-artist | -0.14 | 0.02 | 542271.12 | -7.38 | < .001 | 9.77E-05 | 597988 | 80845 | 10455 | 374465 | 0.142 |
| 9/11 | Self-references | Between-artist | 0.18 | 0.03 | 73478.51 | 5.37 | < .001 | 3.77E-04 | 597988 | 80845 | 10455 | 374465 | 0.142 |
| 9/11 | Collective-references | Intercept | 1.09 | 0.05 | 69.91 | 20.16 | < .001 | | 597988 | 80845 | 10455 | 374465 | 0.072 |
| 9/11 | Collective-references | Within-artist | 0.11 | 0.01 | 552273.3 | 13.66 | < .001 | 3.38E-04 | 597988 | 80845 | 10455 | 374465 | 0.072 |
| 9/11 | Collective-references | Between-artist | 0.14 | 0.01 | 62640.27 | 11.33 | < .001 | 2.03E-03 | 597988 | 80845 | 10455 | 374465 | 0.072 |
| 9/11 | Analytic thinking | Intercept | 44.43 | 1.29 | 83.44 | 34.46 | < .001 | | 597988 | 80845 | 10455 | 374465 | 0.228 |
| 9/11 | Analytic thinking | Within-artist | -0.37 | 0.1 | 533736.2 | -3.74 | < .001 | 2.62E-05 | 597988 | 80845 | 10455 | 374465 | 0.228 |
| 9/11 | Analytic thinking | Between-artist | -1.95 | 0.2 | 80351.81 | -9.8 | < .001 | 1.19E-03 | 597988 | 80845 | 10455 | 374465 | 0.228 |
| 9/11 | Cognitive processes | Intercept | 7.44 | 0.21 | 67.3 | 35.95 | < .001 | | 597988 | 80845 | 10455 | 374465 | 0.160 |
| 9/11 | Cognitive processes | Within-artist | 0.04 | 0.02 | 538601.91 | 2.07 | .039 | 8.46E-06 | 597988 | 80845 | 10455 | 374465 | 0.160 |
| 9/11 | Cognitive processes | Between-artist | 0.18 | 0.03 | 72931.15 | 5.52 | < .001 | 3.89E-04 | 597988 | 80845 | 10455 | 374465 | 0.160 |
| COVID-19 | Self-references | Intercept | 7.16 | 0.19 | 65.65 | 37.58 | < .001 | | 1099094 | 245812 | 19824 | 614077 | 0.191 |
| COVID-19 | Self-references | Within-artist | -0.03 | 0.01 | 919760.26 | -1.94 | .053 | 3.17E-06 | 1099094 | 245812 | 19824 | 614077 | 0.191 |
| COVID-19 | Self-references | Between-artist | 0.63 | 0.02 | 240179.02 | 31.68 | < .001 | 4.19E-03 | 1099094 | 245812 | 19824 | 614077 | 0.191 |
| COVID-19 | Collective-references | Intercept | 1.47 | 0.05 | 95.23 | 28.69 | < .001 | | 1099094 | 245812 | 19824 | 614077 | 0.108 |
| COVID-19 | Collective-references | Within-artist | -0.01 | 0.01 | 933468.61 | -1.83 | .067 | 3.29E-06 | 1099094 | 245812 | 19824 | 614077 | 0.108 |
| COVID-19 | Collective-references | Between-artist | -0.16 | 0.01 | 189399.3 | -18.84 | < .001 | 1.85E-03 | 1099094 | 245812 | 19824 | 614077 | 0.108 |
| COVID-19 | Analytic thinking | Intercept | 41.89 | 1.16 | 84.03 | 36.09 | < .001 | | 1099094 | 245812 | 19824 | 614077 | 0.268 |
| COVID-19 | Analytic thinking | Within-artist | 0.51 | 0.07 | 902912.33 | 6.93 | < .001 | 5.32E-05 | 1099094 | 245812 | 19824 | 614077 | 0.268 |
| COVID-19 | Analytic thinking | Between-artist | -1.77 | 0.11 | 259132.72 | -15.67 | < .001 | 9.47E-04 | 1099094 | 245812 | 19824 | 614077 | 0.268 |
| COVID-19 | Cognitive processes | Intercept | 7.58 | 0.19 | 83.98 | 39.01 | < .001 | | 1099094 | 245812 | 19824 | 614077 | 0.187 |
| COVID-19 | Cognitive processes | Within-artist | 0.12 | 0.01 | 911549.75 | 7.92 | < .001 | 7.12E-05 | 1099094 | 245812 | 19824 | 614077 | 0.187 |
| COVID-19 | Cognitive processes | Between-artist | 0.35 | 0.02 | 228168.45 | 17.23 | < .001 | 1.33E-03 | 1099094 | 245812 | 19824 | 614077 | 0.187 |

*Note*. Observations = number of songs used in the model; Artists = the number of different artists the songs originated from; Spanning = how many artists released enough songs on both sides of the upheaval (at least two each) to be counted; Informative = how many songs can speak to change within-artist. ICC = Intraclass correlation coefficient.

**Table 3**

*Themes Extracted Using the Meaning Extraction Method*

| C1 | | C2 | | C3 | | C4 | | C5 | | C6 | | C7 | |
|---|---|---|---|---|---|---|---|---|---|---|---|---|---|
| Status and hustle | | Courtship | | Light-dark imagery | | Devotion | | Moral conflict | | Recounting | | Life and freedom | |
| λ | % | λ | % | λ | % | λ | % | λ | % | λ | % | λ | % |
| 2.12 | 1.75 | 1.63 | 1.34 | 1.43 | 1.18 | 1.37 | 1.13 | 1.35 | 1.12 | 1.35 | 1.11 | 1.34 | 1.11 |
| Word | Loading | Word | Loading | Word | Loading | Word | Loading | Word | Loading | Word | Loading | Word | Loading |
| money | 0.504 | baby | 0.582 | light | 0.548 | heart | 0.565 | wrong | 0.582 | knew | 0.566 | living | 0.573 |
| big | 0.449 | girl | 0.573 | dark | 0.544 | love | 0.508 | right | 0.541 | thought | 0.509 | live | 0.538 |
| real | 0.401 | wanna | 0.347 | fire | 0.310 | soul | 0.313 | fight | 0.311 | said | 0.475 | life | 0.501 |
| gotta | 0.348 | crazy | 0.301 | black | 0.299 | give | 0.252 | side | 0.286 | told | 0.417 | free | 0.229 |
| boy | 0.330 | tonight | 0.237 | | | | | | | | | | |
| man | 0.328 | | | | | | | | | | | | |
| play | 0.306 | | | | | | | | | | | | |
| watch | 0.231 | | | | | | | | | | | | |
| better | 0.225 | | | | | | | | | | | | |
| hard | 0.224 | | | | | | | | | | | | |

*Note*. Up to 10 terms are represented per theme. λ = eigenvalue. % = amount of variance explained for each component. C1-C7 = components 1 through 7.

**Figure 1**

*Descriptive Trends in Language Dimensions Over Time in 3 Million Music Lyrics*

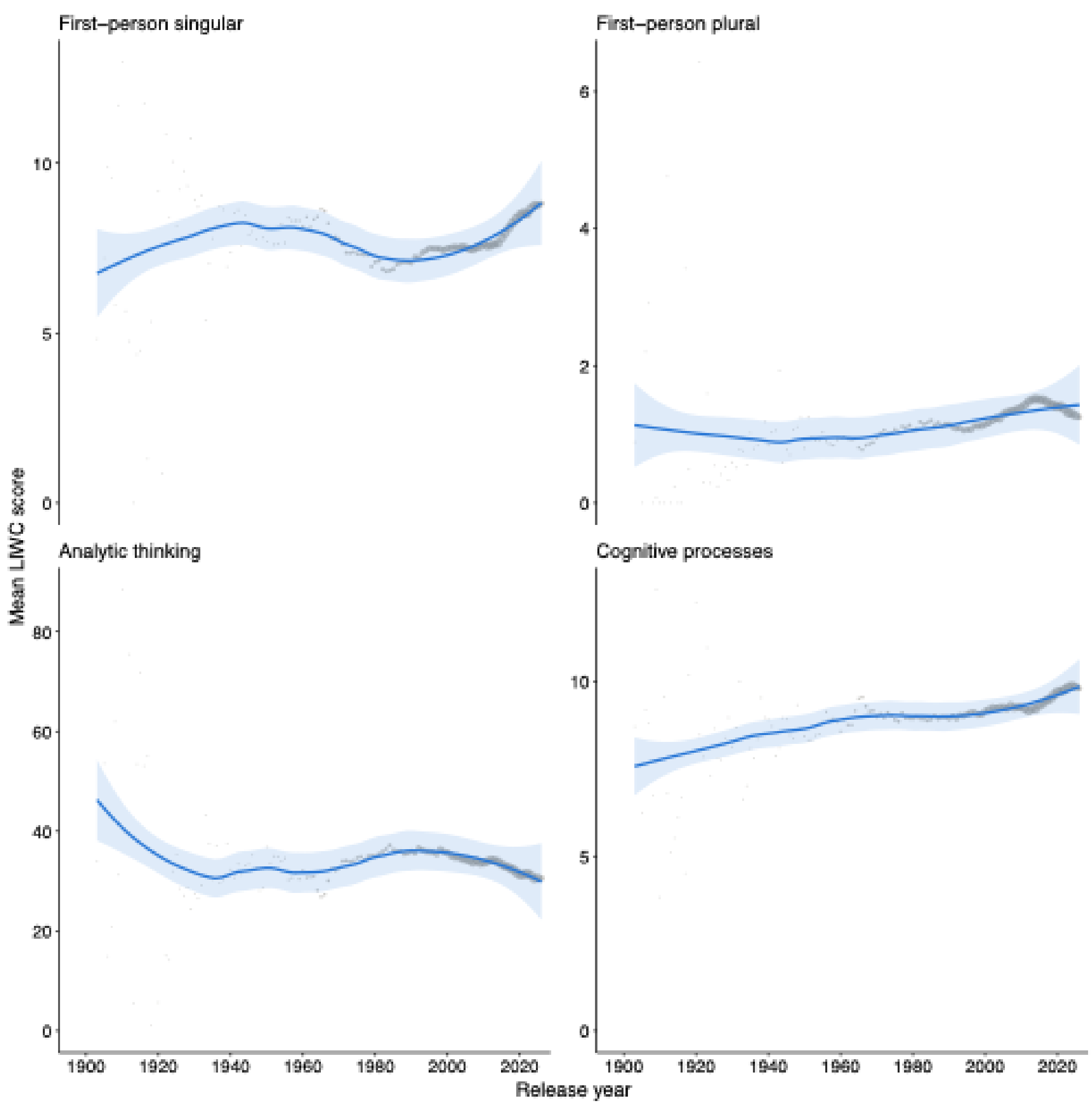

*Note*. Dots in each figure correspond to year-specific means.

**Figure 2**

*Within-Artist Deviations from Baseline for Cultural Upheavals*

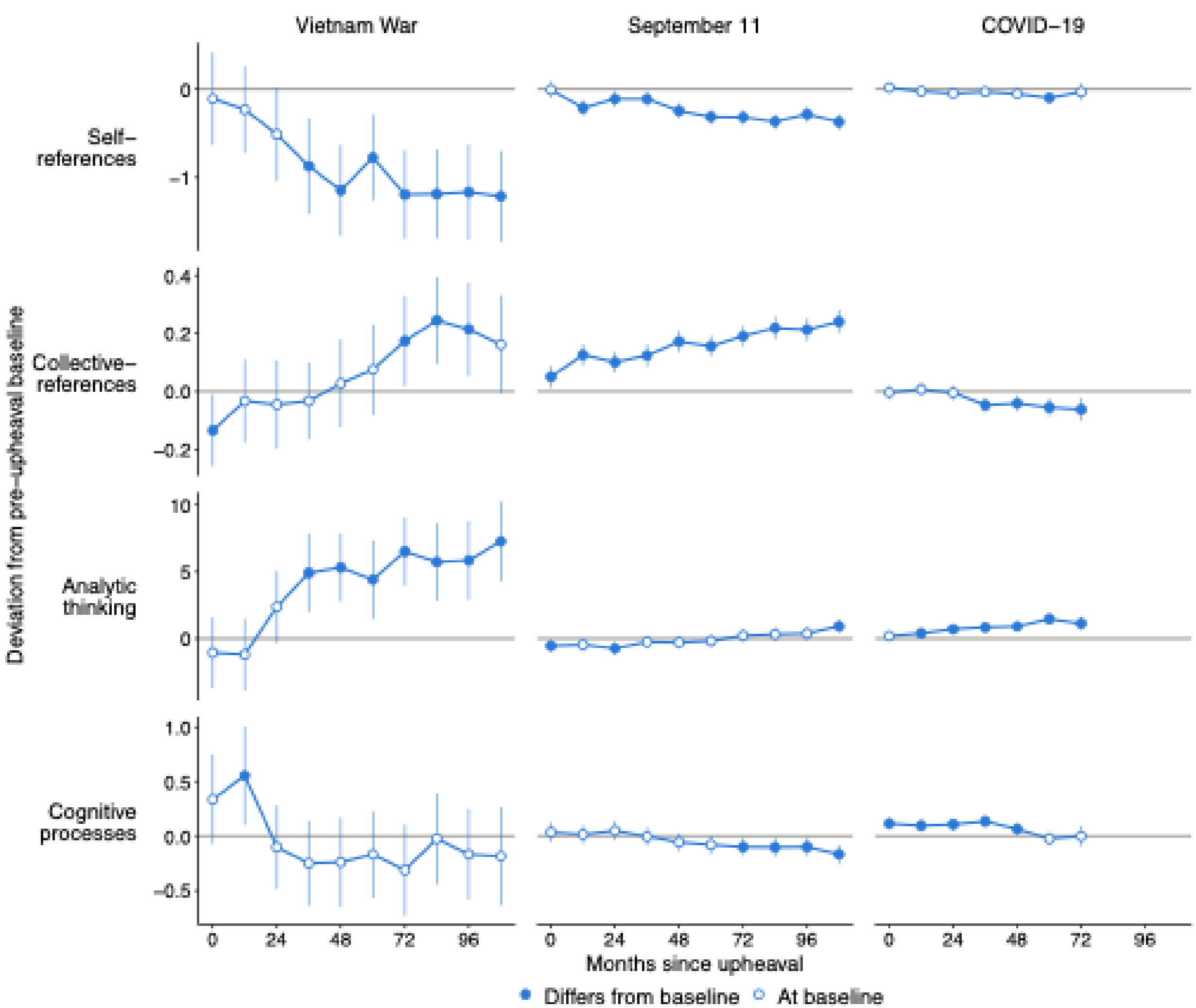

*Note*. Error bars are 95% Confidence Intervals. Points represent the mean deviation from the pre-upheaval baseline within successive 12-month bins plotted at the start of each bin. For example, the point at 0 covers months 0 through 12. The baseline is the pooled pre-upheaval period, which serves as the reference category and is therefore not plotted. Bin-level estimates derive from fixed-effects models and are not directly comparable to the pooled random-intercept estimates in Table 2.